\documentclass[11pt]{article}
\usepackage[preprint]{acl}
\usepackage{times}
\usepackage{latexsym}
\usepackage[T1]{fontenc}
\usepackage[utf8]{inputenc}
\usepackage{microtype}
\usepackage{graphicx}
\usepackage{booktabs}

\title{Pronunciation-Supervised Text-to-Speech for Irish (Gaeilge)}

\author{William Lane \\
  Laneguage \\
  \texttt{will@laneguage.com}}

\begin{document}
\maketitle

\begin{abstract}
Speech technology for Irish must be built from very little
transcribed speech, yet the language's pronunciation is thoroughly
documented in dictionaries and dialect descriptions. We ask whether
that documentation can supervise the adaptation of a large
pre-trained text-to-speech model to a language it was never trained
on. We adapt Qwen3-TTS to Irish using 12.8 hours of crowdsourced
Common Voice speech, correcting two defects in the vendor
fine-tuning script that training loss does not reveal
(Appendix~\ref{sec:appendix-diag}), and reach character error rate
(CER) 0.225 on held-out cross-corpus text, against 0.397 for an
exhaustively tuned from-scratch system and 0.437 for zero-shot
cloning. We then compile the Wiktionary pronunciation record into a
lexicon and a phoneme critic, and use that critic as the verifier in
one round of rejection fine-tuning: CER 0.159 cross-corpus and 0.067
in-domain, with the voice still anonymous across all 202 training
speakers. The same critic explains why the voice cannot yet be
steered to a minority dialect: 229 of 12,311 training clips lean
Ulster against 8,451 Munster. Pronunciation documentation,
converted into a verifier, is a practical lever for low-resource
speech synthesis, and a way to measure what the available data
cannot support.
\end{abstract}

\section{Introduction}
\label{sec:intro}

Irish is an official language of Ireland and the European Union,
with three major dialects (Ulster, Connacht and Munster), a deep
orthography (spelling that does not determine sound), and
productive initial mutations (lenition and eclipsis, which change a
word's opening consonant in grammatical context). These properties make
pronunciation central to Irish speech technology: a system that
reads the letters without the sound system behind them will
mispronounce a substantial share of ordinary words. At the same
time, transcribed Irish speech is scarce. The largest open corpus,
Common Voice, provides some 13 hours of validated read speech, much
of it recorded by learners; the reference Irish synthesis
programme, ABAIR at Trinity College Dublin, rests on roughly 40
hours of project-collected studio and field recordings that are not
public artefacts \cite{murphy2023abair}.

Large pre-trained text-to-speech models have changed what a small
team can reasonably attempt, but they cover only their training
languages, and the strongest open systems either exclude Irish or
carry non-commercial terms. This leaves a practical question with a
methodological core: \emph{how should a pre-trained model be
adapted to an unsupported language when speech is scarce but
pronunciation is documented?} Irish is a good test case precisely
because the imbalance is so clear: a dictionary tradition with
dialect-tagged pronunciations, against a handful of hours of usable
speech.

The methodological claim can be stated in one line: \emph{the
dictionary is the reward model}. Where alignment methods for
speech generation ordinarily require human preferences or a
strong ASR system in the target language, many low-resource
languages possess neither, yet they very often possess a
pronunciation dictionary. We treat that static document as a
verifier: compiled into a lexicon and a phoneme critic, it
scores synthetic speech at the level of phones, selects
training samples, and audits the corpus itself. We are careful
about what that substitution buys: as a selector it matches,
rather than beats, a strong ASR-based selector, and its distinctive
contributions are that it measures pronunciation and dialect
directly and that it needs no target-language ASR of record. The
remainder of the paper tests how far the substitution carries.

We make four contributions. (1)~Evidence that a large
pre-trained TTS model can be adapted to a language absent from
its pre-training with 12.8 hours of crowdsourced speech, into a
single anonymous voice that outperforms both an exhaustively
tuned from-scratch system and zero-shot voice cloning (the
adaptation required two corrections to the vendor fine-tuning
script, documented in Appendix~\ref{sec:appendix-diag}).
(2)~A population voice construction (all training speakers
under one voice tag) that passes a nearest-speaker anonymity
test where small-donor blending measurably fails.
(3)~A pronunciation critic built entirely from static public
resources (a Wiktionary-derived, dialect-tagged lexicon and
lexicon-labelled speech), validated as a measurement instrument,
and used as the verifier in rejection fine-tuning.
(4)~A measured account of dialect in the available data: the
critic mines the corpus's dialect distribution (69\%
Munster-leaning; 229 Ulster-leaning clips), which explains why
minority-dialect steering fails by every route we tried. Every claim we make is a machine
measurement; native-listener evaluation, the remaining gate for
naturalness and dialect authenticity, has not been run, and
Section~\ref{sec:eval} states what that leaves open.

\section{Related Work}

ABAIR's synthesis line evolved from concatenative and
statistical-parametric systems to multi-dialect FastPitch with
HiFi-GAN, trained on project-collected recordings with dialect
letter-to-sound rules \cite{murphy2023abair}. Their ASR work
reports dialect-stratified error rates and warns that crowdsourced
Irish is predominantly non-native \cite{lonergan2024fotheidil,
lonergan2022eist}; BlasBench now provides an open Irish ASR
benchmark, whose leading system we use as an independent scorer
\cite{blasbench2026}.

\paragraph{Adapting large TTS models.} Among large pre-trained TTS
systems, none supports Irish, and XTTS-v2 and MMS carry
non-commercial terms \cite{xtts2024,mms2023}. Qwen3-TTS is, to our
knowledge, the only current frontier-class system released under
Apache-2.0 \cite{qwen3tts}. Fine-tuning such models for
lower-resource languages is an active area: LoRA adaptation of
VoxCPM2 has been reported for Khmer and Korean \cite{voxcpm2026},
and adapter-based adaptation to unseen languages has been studied
for lightweight TTS \cite{adapters2025}. Our setting differs in
three respects: the target language is absent from the base
model's pre-training languages (not merely low-resource); the
adapted voice must be anonymous rather than a cloned or
single-speaker voice; and every comparison is powered and
replicated under a frozen protocol (\S\ref{sec:eval}).

\paragraph{Pronunciation knowledge and assessment.} Dictionary
knowledge has previously been injected into end-to-end TTS at
scale \cite{dict-tts}, and phone-versus-grapheme inputs have been
compared for low-resource transfer \cite{dijkstra2023}.
Pronunciation scoring itself descends from goodness-of-pronunciation
in computer-assisted pronunciation training \cite{gop2000}. We
use that lineage differently: the scorer is trained from
dictionary pronunciations alone, validated as an instrument, and
then used as a data-selection verifier and corpus-mining tool.

\paragraph{Preference and rejection methods for TTS.} Rejection
fine-tuning (best-of-$n$ selection followed by supervised
training) originates in language-model alignment \cite{raft2023}.
For speech, SpeechAlign aligns codec language models to human
preferences with iterative preference optimisation
\cite{speechalign2024}, INTP applies DPO with an
intelligibility-preference dataset \cite{intp2025}, and recent
analyses of fine-tuning failure in LLM-based TTS emphasise data
diversity over naive supervised updates \cite{ftfails2026}. Our
verifier is neither human preference nor ASR agreement: it is a
dictionary-derived phoneme critic, and we compare it directly
against ASR-CER selection under matched conditions
(\S\ref{sec:raft}).

\paragraph{Voice anonymisation.} Speaker anonymisation is
usually framed as post-hoc transformation of speaker embeddings,
classically averaging or resampling x-vectors under the
VoicePrivacy evaluation framework \cite{voiceprivacy2020,
panariello2024}. Our construction differs: anonymity is obtained
at training time by pooling all speakers under one voice tag,
and evaluated by a nearest-speaker test over the training
population (\S\ref{sec:anon}). We are not aware of prior work
that evaluates small-donor \emph{blending} this way, nor of
dictionary-derived dialect pronunciations used as a verifier
for sample selection.

\section{Data and Rights}

All components are recorded in a rights ledger with three states (commercial-cleared, research-only,
blocked-pending-counsel), and product-path runs fail closed on uncleared components.
Training speech is Common Voice Irish (CC0): after hard checks,
12,311 clips from 202 speakers, 12.84 hours \cite{commonvoice}.
FLEURS and FLEURS-R Irish (CC BY 4.0) are held out for evaluation
\cite{fleurs,fleursr}. The pronunciation lexicon derives from the
English Wiktionary dump via kaikki.org (CC BY-SA), acquired with
dump date and hash recorded. Self-reported accent labels exist for
21\% of Common Voice clips and are used only as weak validation
signal; no dialect labels exist for the training speakers, which
is the motivation for the pronunciation-based mining in
Section~\ref{sec:dialect}.

\section{The From-Scratch Baseline}

We first trained Matcha-TTS \cite{matcha} with a HiFi-GAN vocoder
\cite{hifigan}, both from scratch, under a frozen evaluation
protocol (Section~\ref{sec:eval}). The campaign was extensive (step scaling, augmentation, oversampling, checkpoint averaging,
inference variants and an additional corpus ingest were each measured), and it converges to a clear conclusion: at this data
scale the ceiling is set by the data, not by tuning. The best
from-scratch system reaches CER 0.397 on held-out cross-corpus
text (FLEURS-223; ground-truth floor 0.096) and the authors judged it
unnatural in informal listening. Run-to-run seed variance
(0.013--0.027 CER) is of the same order as most intervention
effects, a finding that shapes the replication requirements in
Section~\ref{sec:eval}. We report this phase briefly because its
role is to establish the baseline that adaptation must beat, and
because its negative results motivate the change of approach.

\section{Adapting a Pre-trained Model to an Unseen Language}

\subsection{Zero-shot transfer}

Qwen3-TTS 1.7B, prompted with a single Irish reference clip,
synthesises a language absent from its ten pre-training languages:
on the powered comparison of Section~\ref{sec:eval} it scores CER
0.437 cross-corpus, with the highest quality proxy (DNSMOS OVRL
3.42) of any early system. It does so by cloning the reference
speaker, however, which our voice-identity posture excludes from
any product path; it also remains unstable on short prompts (2 of
277 evaluation clips ran to the token cap). We treat zero-shot
cloning as a research baseline and a measure of what pre-training
already knows about Irish phonology.

\subsection{Adaptation recipe}

\begin{figure}[t]
\centering
\includegraphics[width=\columnwidth]{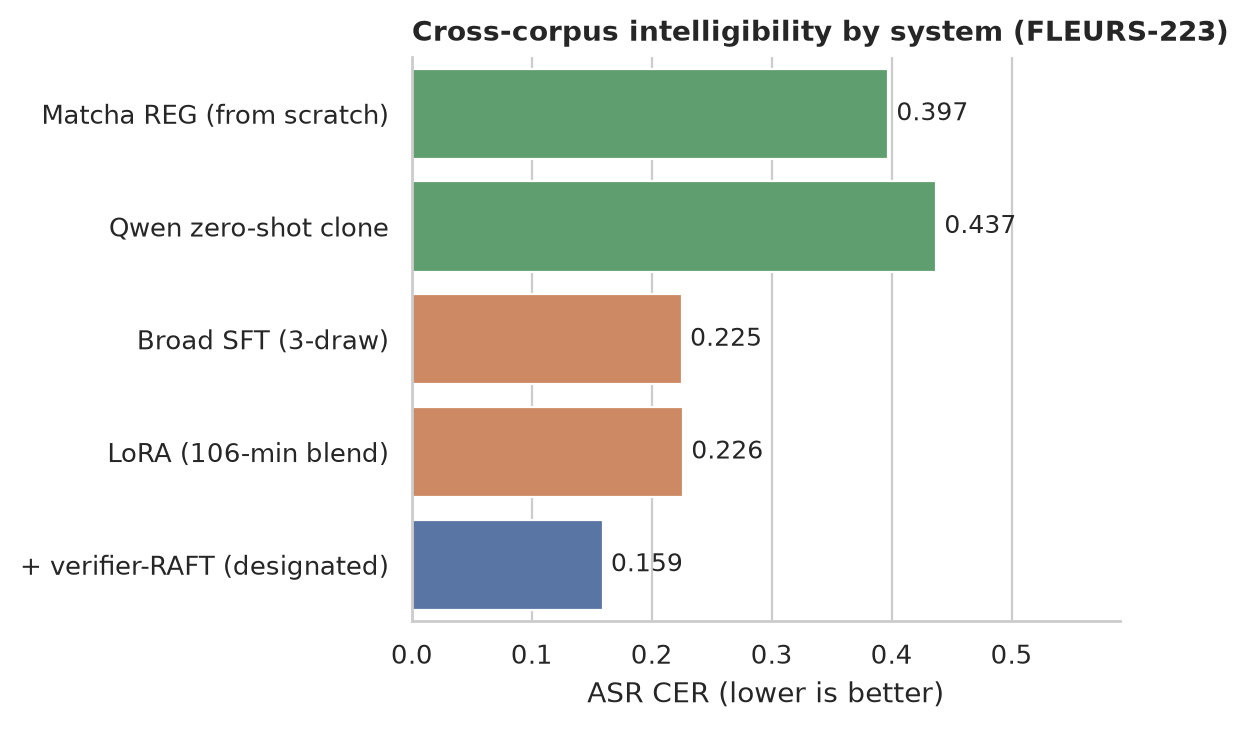}
\caption{Cross-corpus intelligibility by system. Adaptation, not
further tuning of the from-scratch recipe, closes the gap.}
\label{fig:systems}
\end{figure}

Our target is a single anonymous voice, so we fine-tune the 0.6B
base model under one voice tag, with the speaker slot set to the
mean of donor reference embeddings (provenance recorded). We call
the resulting voice \emph{Gl\'or Nua} (``new voice''); its release
is described with the other artifacts in
Section~\ref{sec:discussion}. The
vendor fine-tuning script required two corrections before any
run could be trusted (a double label shift, and a sub-talker
objective whose input already contains its target codes), together
with a lower learning rate (1e-5 rather than the
default 2e-5); we verified the corrections with a
teacher-forced alignment probe before training (details and
probe measurements in Appendix~\ref{sec:appendix-diag}). These
are implementation matters, reported for reproducibility; no
claim in what follows depends on them beyond the fact that the
adapted models below were trained with the corrected recipe. At
that recipe, the adapted models cross over the baseline
decisively (Table~\ref{tab:main}, Figure~\ref{fig:systems}).
Training on the full 202-speaker population under one voice tag
outperforms the quality-gated three-donor blend at both model
scales, and rank-16 LoRA adapters on the 106-minute blend match
the 12.8-hour full fine-tune cross-corpus: on small, clean

data, adapter regularisation substitutes for roughly seven times
the data volume. Pure greedy decoding collapses to silence at the
token cap; the vendor sampling defaults stand.

\begin{table}[t]
\centering
\footnotesize
\setlength{\tabcolsep}{3pt}
\begin{tabular*}{\columnwidth}{@{\extracolsep{\fill}}lrrr}
\toprule
System & common-80 & FLEURS-223 & Anon. \\
\midrule
Matcha REG & 0.216 & 0.397 & fixed \\
Qwen zero-shot clone & 0.317 & 0.437 & clone \\
Broad SFT (3-draw) & 0.102 & 0.225 & PASS \\
LoRA, 106 min blend & 0.117 & 0.226 & FAIL \\
\textbf{+ verifier-RAFT} & \textbf{0.067} & \textbf{0.159} & \textbf{PASS} \\
\bottomrule
\end{tabular*}
\caption{Final systems (ASR CER, lower is better; three-draw
means where noted; ``verifier-RAFT'' is rejection fine-tuning,
Section~\ref{sec:raft}). Ground-truth floors: common-80 0.054,
FLEURS-223 0.096 (same scorer on real speech). Anonymity: donor /
nearest-speaker similarity against its bar.}
\label{tab:main}
\end{table}

\subsection{Voice anonymity}
\label{sec:anon}

A blended voice is not automatically an anonymous one. Our
three-donor blends, pitch-shifted and pooled under one tag, sit at
similarity 0.88--0.90 to their nearest donor against a gate bar of
0.82 (donor self-similarity 0.94--0.98): the voice collapses
toward one donor, and similarity \emph{rises} as the model learns
the donors' Irish better. Training instead on the whole
202-speaker population under one tag passes a nearest-speaker
generalisation of the same test (0.87--0.88 against a bar of
0.93, over centroids from every training speaker). The two tests use
different bars because their comparison sets differ: the donor test
asks whether a voice is unusually close to one of three donors, the
population test whether it approaches the population's maximum
inter-speaker similarity (0.962). Anonymity, in
this setting, is a property of population averaging, not of
blending a few voices. Both tests use a single speaker-embedding
model and a single protocol, so this result is evidence rather than
a guarantee.

\section{The Dictionary as Verifier: Lexicon and Critic}

The lexicon is built from the Wiktionary dump without any
institutional cooperation: 11,532 words with IPA transcriptions,
dialect-tagged for Ulster (9,253 transcriptions), Munster (6,545)
and Connacht (5,473, including subdialect tags mapped to their
major group). Of these, 4,354 word types carry different
pronunciations across dialect groups, and such dialect-diagnostic
words make up 33\% of running tokens in our corpora; overall token
coverage is 82\%. A small dialect-conditioned grapheme-to-phoneme
model trained on the lexicon reaches only 0.51 phone error rate
(PER) on held-out words: the source transcriptions are too variable for it to be
a label source, so it is used as an auxiliary only. Silver
phone labels for the full training corpus come directly from the
lexicon (out-of-vocabulary words, 15\% of tokens, are omitted; the
diagnostic vocabulary is in-lexicon by construction).

\begin{figure}[t]
\centering
\includegraphics[width=\columnwidth]{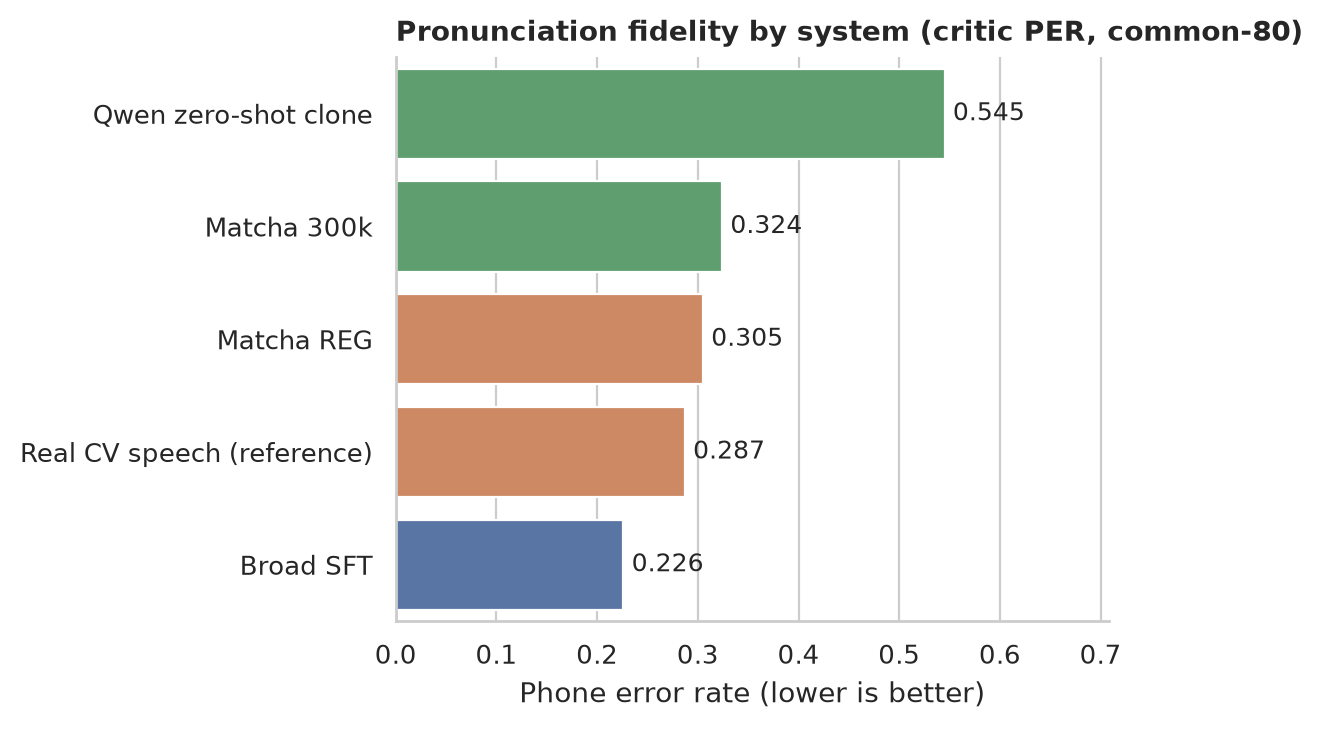}
\caption{Pronunciation fidelity as heard by the critic.
ASR-based scoring alone does not separate these systems this way.}
\label{fig:critic}
\end{figure}

\begin{figure*}[t]
\centering
\includegraphics[width=0.92\textwidth]{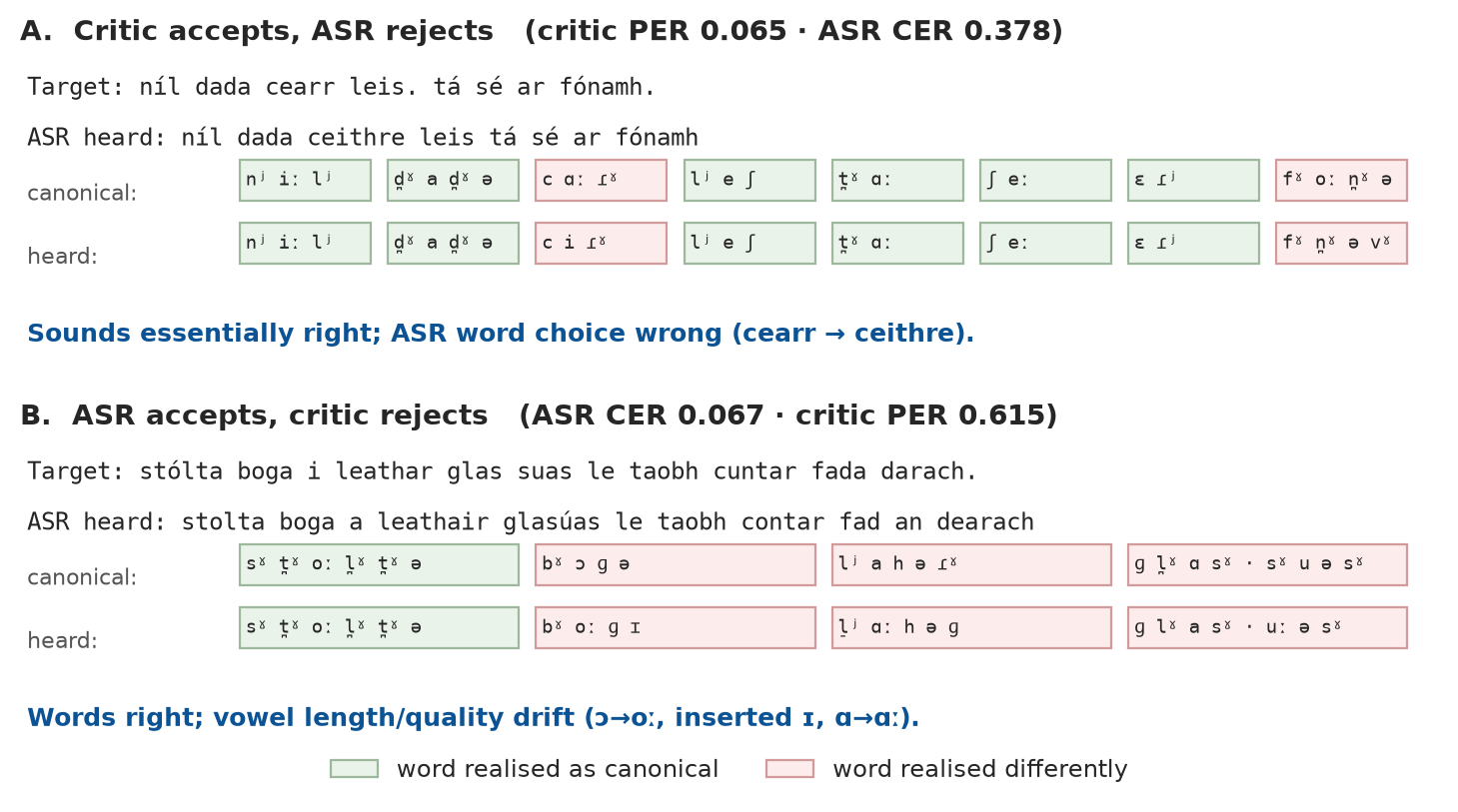}
\caption{Two clips the two scorers judge oppositely (from the
1,000-sample pool). (A) The pronunciation is essentially
canonical, but the ASR substitutes a real word
(\emph{cearr} $\rightarrow$ \emph{ceithre}), so CER counts a
pronunciation success as failure. (B) The ASR transcript is
near-perfect while vowel length and quality drift from the
canonical phones, which CER cannot see. This asymmetry is the
scientific reason for a pronunciation-level verifier.}
\label{fig:specimen}
\end{figure*}

The critic is a wav2vec2-XLS-R (300M) CTC phoneme recogniser
fine-tuned on those silver labels \cite{xlsr}. It reaches
held-out phone error rate 0.164 on speakers it never saw in
training. As a system scorer it orders our checkpoints by PER on
common-80 (Broad SFT 0.226 $<$ Matcha REG 0.305 $<$ Matcha 300k
0.324 $<$ zero-shot clone 0.545, on a scale that is not comparable
to the CER numbers of Table~\ref{tab:main}), consistent with the
CER ordering, but not
redundant with it: the clone, which the quality proxy ranks highest,
pronounces worst (Figures~\ref{fig:critic} and
\ref{fig:specimen}). One validation
result determines how the critic may be used. Scored against a
single canonical pronunciation, native speakers from the Gaeltacht
(Irish-speaking regions) in held-out data score PER 0.61 against
0.23 for unlabeled speakers.
The critic therefore measures conformity to a standard-leaning
canonical, not nativeness; dialect speakers deviate from it by
design. Any reward or filter built on the critic must use
dialect-matched canonicals, and nativeness-band PER is rejected
as a validation gate. This is the kind of instrument error that
is cheap to find before the critic is trusted with selection, and
expensive afterwards.

\section{Verifier-Guided Fine-tuning from Dictionary Rewards}
\label{sec:raft}

Rejection fine-tuning (RAFT) is the simplest way to let a
verifier improve a generator: sample candidates, keep those the
verifier scores best, and fine-tune on them. From the broad SFT
checkpoint we sampled four candidates for each of 250 prompts at
temperature 1.0, scored each with the critic (phone error rate
against the canonical pronunciation, a dialect margin term, and
an ASR-CER guard against unintelligible samples), kept the best
sample per prompt where it cleared a margin floor
(84 of 1,000 clips) and trained a rank-16 LoRA adapter on them under the
same single voice. One such round moves FLEURS-223 CER from 0.222
to 0.153 and in-domain CER from 0.090 to 0.059, with the quality
proxy unchanged and anonymity intact (single draw; the
three-draw means behind the headline comparison are 0.225 and
0.159, Section~\ref{sec:eval}). A second round changes
nothing (paired difference not significant): the model's own
samples are already near the verifier's ceiling, so there is
little left to select. The mechanism is worth stating plainly: temperature sampling produces variation, the verifier finds the
unusually clean cases, and training moves the model's average
output toward its own best behaviour.

\begin{figure}[t]
\centering
\includegraphics[width=\columnwidth]{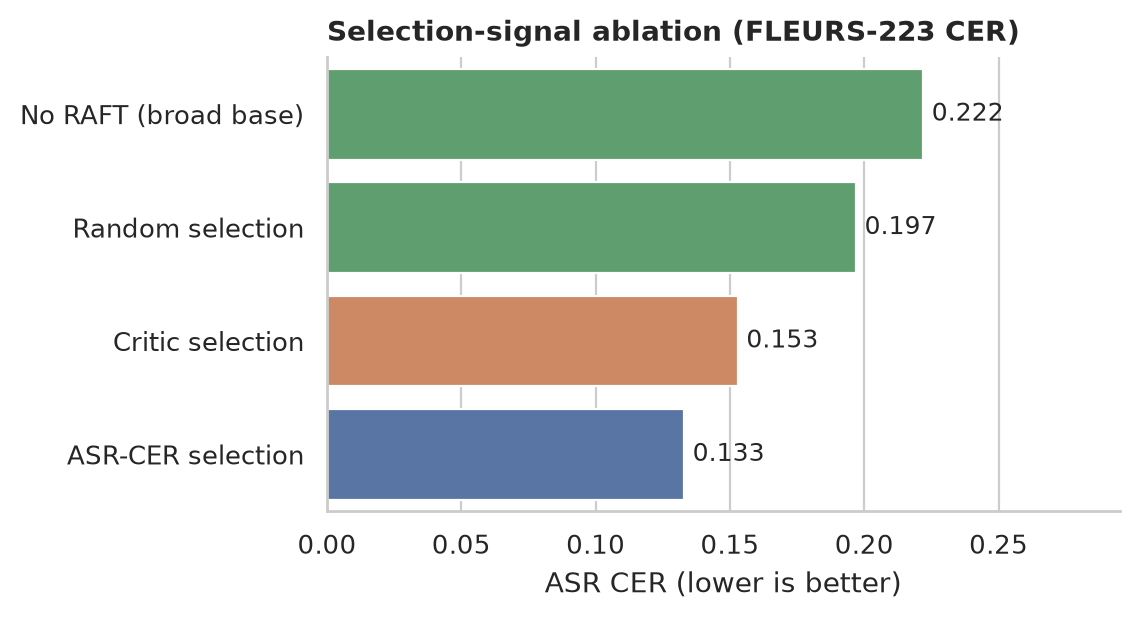}
\caption{Selection-signal ablation (matched 84-clip arms from one
sample pool). Selection matters more than which strong signal
selects.}
\label{fig:ablation}
\end{figure}

\begin{figure}[t]
\centering
\includegraphics[width=\columnwidth]{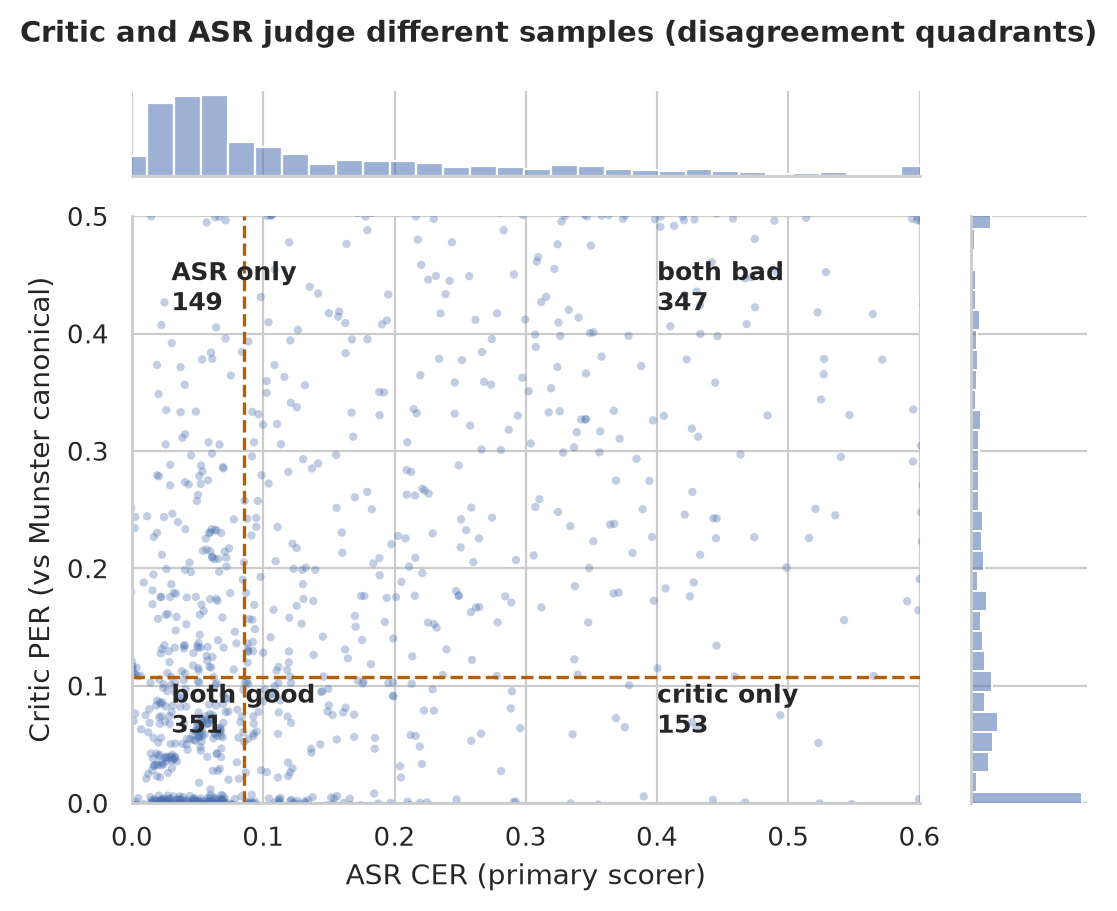}
\caption{Joint behaviour of the two selection signals over the
1,000 sampled clips. The quadrants at the medians are almost
balanced (both-good 351, both-bad 347, critic-only 153,
ASR-only 149): the signals fail on largely different clips,
which is why each can select a winning arm and why neither
dominates under the other's metric.}
\label{fig:joint}
\end{figure}

What does the critic contribute over the obvious alternative,
selecting by ASR CER? The two signals are nearly independent over
the sample pool (correlation 0.097; they agree on the best of
four samples in only 137 of 250 prompts; Figure~\ref{fig:joint}
shows where each fails). Retraining identical
arms on matched 84-clip selections gives: no RAFT 0.222, random
selection 0.197, critic selection 0.153, CER selection 0.133
(FLEURS-223; Figure~\ref{fig:ablation}). Selection itself is
therefore the large effect, and CER selection's edge on CER
endpoints is partly by construction, since it optimises the
metric family that grades it. Under the independent Blas-IE
scorer the two selections tie cross-corpus (0.172 vs 0.175,
difference not significant), while critic selection retains the
best phone fidelity (PER 0.056 vs 0.059; random 0.081). We do not
claim the critic is the better \emph{general} selector. Its
contribution is that it measures pronunciation directly,
independently of any ASR family, and that the same instrument
supports the dialect analysis of the next section, which no CER
signal can perform.

\section{The Dialect Question}
\label{sec:dialect}

\begin{figure}[t]
\centering
\includegraphics[width=\columnwidth]{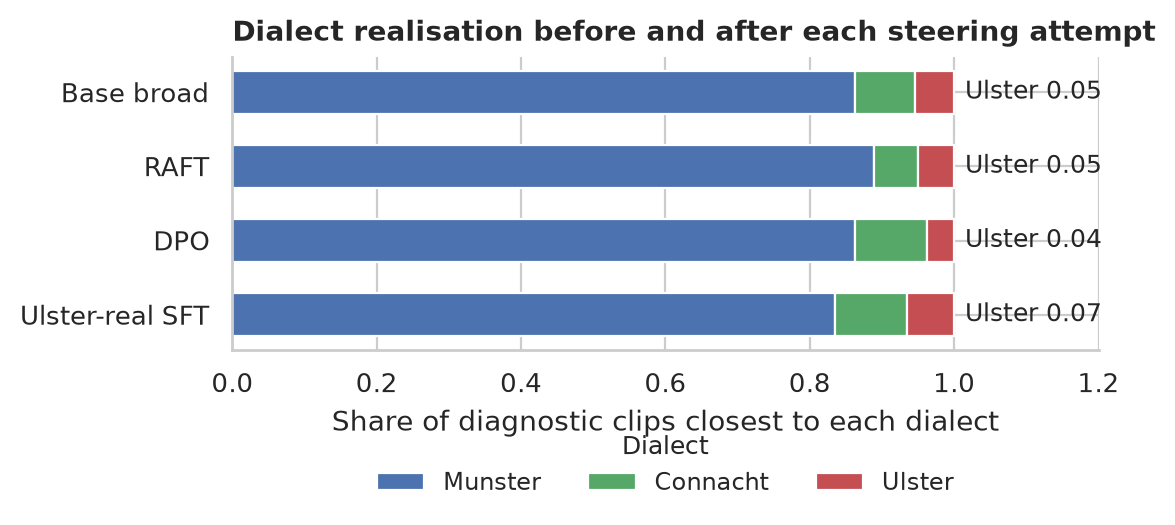}
\caption{Dialect realisation of the voice before and after each
steering attempt. The Munster share never falls.}
\label{fig:dialect}
\end{figure}

\begin{figure}[t]
\centering
\includegraphics[width=\columnwidth]{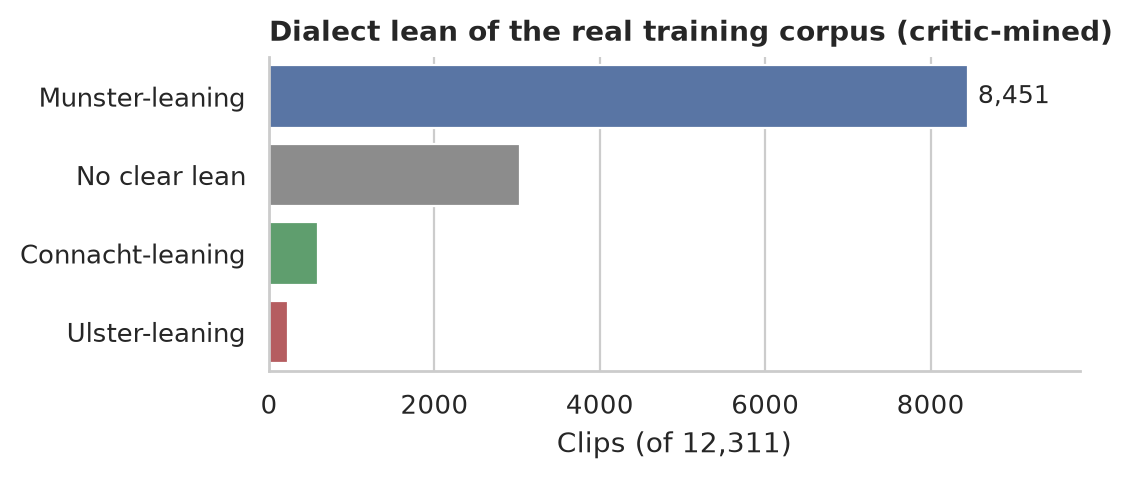}
\caption{Dialect lean of the real training corpus, as mined by
the critic. Figure~\ref{fig:dialect} is explained by this
distribution.}
\label{fig:mining}
\end{figure}

Can the same machinery give the voice a chosen regional accent?
The base voice is already Munster-dominant on held-out diagnostic
sentences: 86\% of clips are pronounced closest to the Munster
canonical, 5\% to Ulster. RAFT raises the Munster share to 89%: rejection fine-tuning on majority-dialect samples is a
majority-dialect amplifier. Preference optimisation on
verifier-ranked margin pairs (DPO) leaves dialect unchanged, and
for a measurable reason: only 25 of the 1,000 policy samples
realise Ulster-leaning phones at all, and the chosen/rejected
pairs differ mainly in clarity, so there is no systematic dialect
contrast in the pair signal for any optimiser to learn. Training
adapters on the real Ulster-leaning clips the critic can find
degrades the voice without steering it. The critic's mining of
the full corpus explains the pattern (Figure~\ref{fig:mining}):
8,451 of 12,311 clips lean Munster, 593 Connacht, and 229 Ulster
(185 strongly, across 61 speakers), mirroring the dialect bias
reported for Irish ASR \cite{lonergan2024fotheidil}.
Minority-dialect synthesis from this resource is not an
optimisation problem awaiting a better algorithm; it is short of
the majority dialect by a factor of about forty in raw clips and by
more still in the samples the model actually produces, and the
verifier is what makes that shortage measurable rather than
anecdotal
(Figure~\ref{fig:dialect}).

\begin{figure*}[t]
\centering
\includegraphics[width=0.92\textwidth]{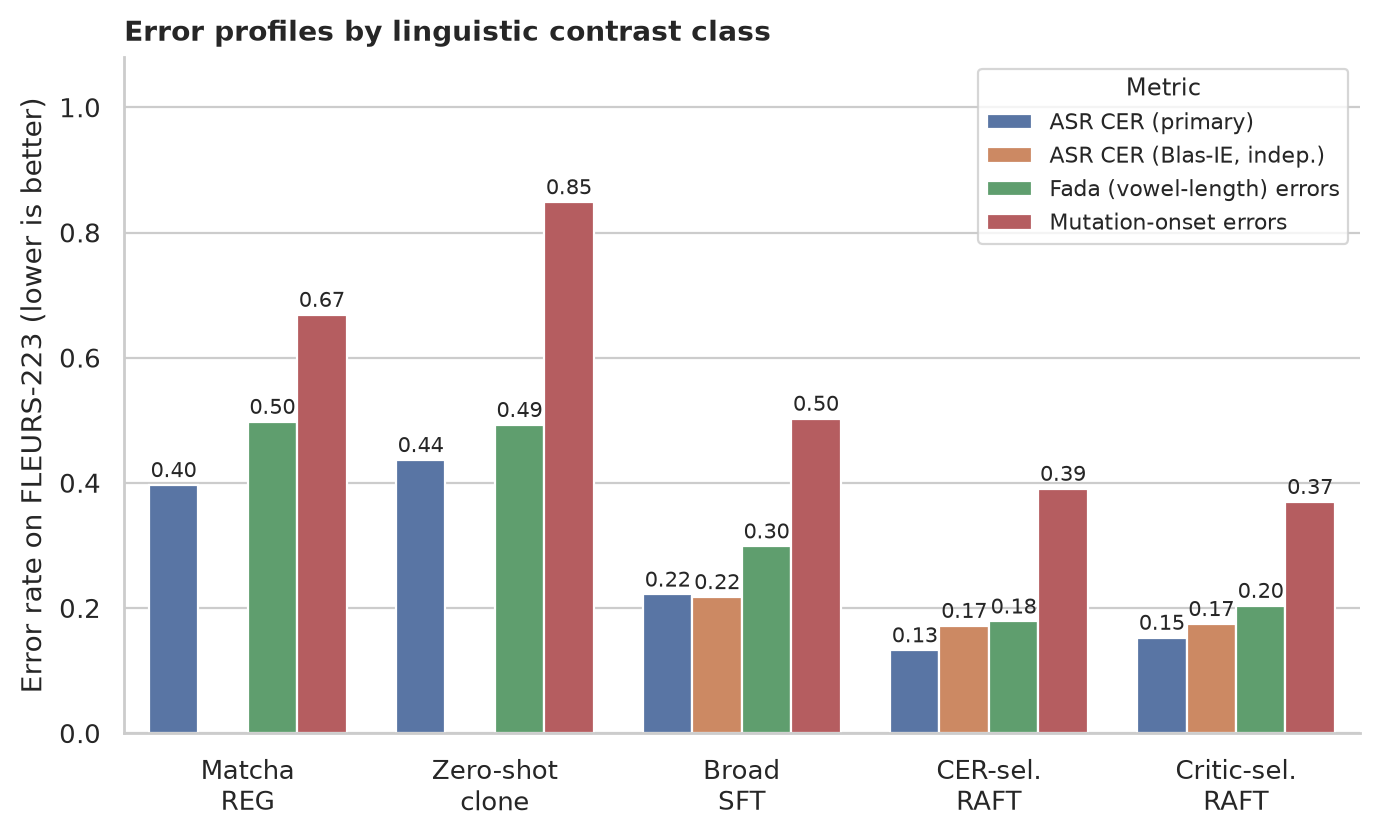}
\caption{Error profiles by linguistic contrast class
(FLEURS-223). Adaptation and verifier fine-tuning reduce not
only overall CER but specifically fada (vowel-length) and
mutation-onset errors; the zero-shot clone's mutation rate
(0.85) exposes what its quality proxy hides. The two RAFT
selection signals end within noise of each other on every
class.}
\label{fig:contrast}
\end{figure*}

\section{Evaluation Protocol}
\label{sec:eval}

Machine metrics are screening instruments. The claims that
ultimately matter (naturalness, pronunciation acceptability,
dialect authenticity) require native listeners, and that study is
not part of this paper: its design is fixed and waits on
recruitment, with a validated challenge set (29 items covering
fadas, lenition, eclipsis, broad/slender contexts, dialect-sensitive
items, numbers, names and code-switching; inter-rater
$\kappa \geq 0.6$ required). We therefore make no naturalness,
acceptability or dialect-authenticity claim anywhere in this paper.
Until then we hold the
machine evaluation to a strict standard: frozen speaker-disjoint
sets (common-80; FLEURS-76/223; in-domain-400; FLEURS-train-200;
long-form probes; gate phrases) reported with ground-truth
floors; paired bootstrap confidence intervals; and replication
across sampling draws and training seeds before any system is
designated. Under that standard the designated candidate's
advantages are: three-draw FLEURS-223 0.159 vs 0.225 ($\Delta$
$-0.066$ [$-0.076$,$-0.056$]); in-domain-400 0.087 vs 0.123;
FLEURS-train-200 0.178 vs 0.262; a second RAFT seed at 0.161
(difference not significant); no long-form regression
(long-form concatenated 0.254 vs 0.296; natural-long 0.146 vs
0.154); contrast-class error rates fall in step, with fada and
mutation-onset errors dropping furthest for the designated voice
(Figure~\ref{fig:contrast}). Because the primary scorer was itself fine-tuned on
related corpora, ranking and headline wins were cross-checked
with Blas-IE \cite{blasbench2026}, the leading open Irish ASR
system, under greedy CTC decoding with no shared training
pipeline: the ordering and the RAFT win replicate (FLEURS-223
0.175 vs 0.218). One early anonymity ``pass'' in our records,
measured on unintelligible output, is retracted; gates are only
ever reported on intelligible speech.

\section{Limitations and Ethics}

No system here clones an identifiable person: the designated
voice is a population average that passes its anonymity test,
while small-donor blends fail theirs and are not product
candidates. A single flagship voice, if wanted, requires
dedicated recordings with explicit consent, which we treat as
blocked pending legal advice. Dialect representation is the
sharpest limitation. Irish speakers are entitled to systems that
speak their own variety; our results show the open corpus cannot
currently support that for Ulster or Connacht, and we regard
targeted, consented dialect data collection
(not further optimisation) as the correct next investment. The critic has
not been validated against human pronunciation judgments in this
version, so its scores are conformity to the lexicon rather than
evidence about nativeness; it also inherits the limits of its silver
labels (lexicon coverage and transcription variability), and its
canonicals are standard-leaning. We have scoped every claim
accordingly.
Absolute error rates are scorer-dependent and are reported only
as within-pipeline comparisons alongside an independent
cross-check. Finally, Common Voice Irish skews toward learner
speech, and any model trained on it inherits that skew; the
critic measures pronunciation conformity, not nativeness, and we
do not present it as doing more.

\section{Discussion}
\label{sec:discussion}

Table~\ref{tab:rq} collects the answers to the questions posed
in Section~\ref{sec:intro}. Three points deserve emphasis. First, the
adaptation result (RQ1) is not a prompt-engineering artefact:
the fine-tuned voice halves the zero-shot clone's error on
powered sets while satisfying an anonymity constraint the clone
cannot meet. Second, the critic's value (RQ2/RQ3) is
instrumental before it is optimising: it ranks systems by
pronunciation where ASR scoring cannot, it exposes its own
canonical-conformity bias before that bias can contaminate
rewards, and only then does it improve the model as a RAFT
verifier, where it matches, rather than beats, the
metric-optimising CER selector under an independent scorer.
Third, the dialect result (RQ4) should be read as a resource
finding, not a modelling failure: three distinct steering
mechanisms fail for the same measured reason, and the verifier
quantifies the missing resource (229 target-dialect clips;
2.5\% target realisations in policy samples). For other
low-resource languages the transferable sequence is: build the
lexicon first, validate the critic as an instrument, mine the
corpus before promising dialect coverage, and only then spend
optimisation effort.

\begin{table}[t]
\centering
\small
\setlength{\tabcolsep}{4pt}
\begin{tabular}{p{0.30\columnwidth}p{0.62\columnwidth}}
\toprule
Question & Answer (evidence) \\
\midrule
RQ1: Can a pre-trained model be adapted to an unseen language
into an anonymous voice? &
Yes. CER 0.225 vs 0.397 (from scratch) and 0.437 (zero-shot);
anonymity PASS for the population voice, FAIL for blends. \\
RQ2: Does documented pronunciation, as a critic, measure and
improve synthesis? &
Critic PER 0.164 (unseen speakers); ranks systems independently
of ASR; one RAFT round: CER 0.222 $\rightarrow$ 0.153. \\
RQ3: Is the critic a better selection signal than ASR CER or
quality scores? &
Not for general CER (tie under independent scorer); uniquely
enables phone-level measurement, dialect analysis and mining. \\
RQ4: Can the voice be steered to a minority dialect? &
No, by RAFT, DPO or real-subset SFT. Corpus is 69\%
majority-leaning; 229/12,311 target clips. \\
\bottomrule
\end{tabular}
\caption{Research questions and their measured answers.}
\label{tab:rq}
\end{table}

\paragraph{Artifacts and reproducibility.} Available now: the
released voice, its model card, demonstration audio and a generation
script. Accompanying the code repository, subject to the rights
ledger: the pronunciation lexicon and diagnostic word lists (CC
BY-SA, with attribution), the critic and G2P code, the corrected
trainer and evaluation scripts, and per-clip scores for every
reported system. The designated voice is
released as \emph{Gl\'or Nua} (the population voice that passes
the anonymity test), with its model card, demonstration audio
and a generation script, under Apache-2.0 at
\url{https://huggingface.co/Laneguage/glor-nua-irish-tts-0.6b}
\cite{glornua2026}; the small-donor blends that fail the
nearest-speaker test are deliberately not released. Every run
record carries the vendor commit, data-manifest hash, seed and
full configuration.
\section{Conclusion}

Adapting a large pre-trained TTS model to a language it was never
trained on proved tractable with remarkably little speech: a
single anonymous Irish voice at CER 0.16 cross-corpus from 12.8
hours of crowdsourced recordings, outperforming both the
from-scratch recipe and the model's own zero-shot cloning. The
methodological point generalises beyond
Irish. Low-resource languages are often documented better than
they are recorded: pronunciation dictionaries, dialect
descriptions and word lists exist where transcribed speech does
not. Converted into a lexicon and a validated critic, that
documentation becomes a verifier that can select training
samples, score systems at the level of phones, and measure the
dialect composition of a corpus. It improved our voice; it also
told us, with numbers, which improvement the data cannot yet
support. We regard both as progress toward speech technology that
serves the varieties people actually speak.

\bibliography{references}

\appendix

\section{Adaptation diagnostics (implementation note)}
\label{sec:appendix-diag}

Vendor fine-tuning (Qwen3-TTS, commit 022e286) failed identically at
both model scales: training loss fell normally and teacher-forced
end-of-sequence prediction was near-perfect, yet free-running
generation did not stop and produced Irish-like babble. We isolated
the two defects with a teacher-forced alignment probe that scores
argmax accuracy at controlled temporal offsets. (i)~The trainer
pre-shifts codec labels and the Transformers loss shifts again, so
the talker is trained two frames ahead; the same applies to the
sub-talker, and one residual group receives no gradient. The
community documents this defect independently, along with a missing
text-projection call \cite{qwen3tts}. (ii)~Sub-talker supervision
pairs each frame's codes with the hidden state at that frame's own
input position, whose embedding already sums all 16 code-group
embeddings of that frame. Measured on held-out clips under the
pairing that generation actually uses, code-predictor accuracy per
group is 0.05--0.17 for the base model and 0.01--0.06 for
vendor-recipe SFT; under the leaky training pairing the same SFT
scores 0.32--0.53. The corrected trainer supervises the sub-talker
at the generation pairing, computes explicit per-group cross-entropy
including group 15, fixes the speaker slot deterministically, and
defaults to learning rate $2\times10^{-6}$ for the smoke gate.

\section{Training configuration}
\label{sec:appendix-training}

Table~\ref{tab:appendix-train} gives the configuration of the two
runs behind the released voice: the adaptation run, and the
rejection fine-tuning round of Section~\ref{sec:raft}.

\begin{table}[t]
\centering
\footnotesize
\setlength{\tabcolsep}{4pt}
\begin{tabular}{lll}
\toprule
 & Adaptation & Rejection FT \\
\midrule
Initialisation & Qwen3-TTS 0.6B & adapted, epoch 2 \\
Data & 12,311 clips & 84 selected clips \\
Epochs & 3 & 3 \\
Batch & 2 $\times$ 4 accum. & 2 $\times$ 4 accum. \\
Learning rate & 1e-5 & 1e-4 \\
Optimiser & AdamW, wd 0.01 & AdamW, wd 0.01 \\
Precision & bf16 & bf16 \\
Adapter & full model & LoRA $r{=}16$, $\alpha{=}32$ \\
Trainable & all & 11.9M \\
Seed & 42 & 42 (replicate 777) \\
\bottomrule
\end{tabular}
\caption{Training configuration. Both runs use the corrected trainer,
a single voice tag (\texttt{ga\_blend\_m1}) and one RTX PRO 6000
(96\,GB).}
\label{tab:appendix-train}
\end{table}

The speaker embedding is the mean of the donor reference embeddings,
computed before training over the distinct reference clips in the
manifest and recorded in the checkpoint metadata; the rejection round
conditions on that existing slot rather than a new one. Checkpoints
are scored at the end of every epoch, and the epoch-2 checkpoint was
carried forward on cross-corpus character error rate.

\section{Rejection fine-tuning: prompts and reward}
\label{sec:appendix-raft}

Candidates are sampled from the adapted checkpoint with the same
prompt set in every round: 200 sentences from a 1,194-sentence
diagnostic set drawn from the pronunciation knowledge base, which
covers the contrast classes of interest (vowel length and fadas,
lenition and eclipsis onsets, broad and slender contexts), plus 50
sentences sampled at even intervals from the training-corpus
transcripts. Four candidates are drawn per prompt at temperature 1.0
for the talker and the sub-talker, giving 1,000 clips.

Each clip is decoded once by the critic and compared with the
canonical pronunciation of its prompt, giving the reward
\[
\mathrm{reward} \;=\; -\mathrm{PER}_{\mathrm{target}}
  \;+\; 0.5\,\mathrm{margin},
\]
\[
\mathrm{margin} \;=\; \min_{d}\mathrm{PER}_{d}
  \;-\; \mathrm{PER}_{\mathrm{target}}.
\]
where the margin compares the target canonical with the other
dialect canonicals, and a penalty of 1.0 is applied when the ASR
character error rate exceeds 0.35, which rejects the candidate in
practice since rewards are of order $-0.1$. The best candidate per
prompt is kept when its margin clears $-0.05$; 84 of 1,000 clips
survive. The designated round scores against the general canonical
lexicon. The dialect-steering experiments of
Section~\ref{sec:dialect} instead score against a single dialect
canonical; that choice is what puts native speakers at PER 0.61
against 0.23 for unlabelled speakers. Sampling uses seed 1234;
training uses seed 42, replicated at 777.

\section{Decoding configuration}
\label{sec:appendix-decoding}

Evaluation and demonstration use the sampling defaults stored with
the checkpoint: temperature 0.9, top-$k$ 50, top-$p$ 1.0 and
repetition penalty 1.05, matched for the sub-talker, with a
1,024-token cap per utterance. Sampling is not optional: pure greedy
decoding collapses to silence at the cap. Because output is
stochastic, headline numbers are means over three draws with paired
bootstrap intervals, and rare runaways to the cap (at most 2 of 319
clips across the reported sets) are counted rather than discarded.
Audio is 24\,kHz.

\section{Evaluation details}
\label{sec:appendix-eval}

Machine scoring uses two systems that share no training pipeline.
The primary scorer is a fine-tuned Irish Whisper
(\path{Eimhin03/MCV_Fleurs_Combined_Irish_ASR}); it overlaps
our training corpora, so it ranks systems but does not set absolute
values. The independent cross-check is Blas-IE
(\path{jyoutir/blas-m3-lm}), a wav2vec2-XLS-R CTC system
transcribed greedily without a language model \cite{blasbench2026}.
Normalisation and metrics are Irish-aware: fadas and initial
mutations are preserved, and beside character error rate we report
fada and mutation-onset error rates.

All sets are frozen and speaker-disjoint from training. Common-80
and FLEURS-223 are the primary held-out sets, in-domain-400 and
FLEURS-train-200 the powered extensions, the long-form probes
concatenate utterances into paragraph-length input, and the gate set
is 16 fixed phrases that check whether generation terminates at all.
Ground-truth floors from the same scorer on real speech are CER
0.054 on common-80 and 0.096 on FLEURS; absolute values are
scorer-dependent, so only within-scorer comparisons carry weight.

\section{Data and artifacts}
\label{sec:appendix-data}

Training speech is Common Voice Irish 27.0, obtained through the
Mozilla Data Collective (record \texttt{cmu5uww2f0095nq07fx63w1tc},
SHA256 \texttt{21e7f007...62ee1}), licensed CC0: 13,473 validated
clips reduce to 12,311 after hard checks, which excluded 31
pathological-silence clips. FLEURS and FLEURS-R Irish (CC BY 4.0)
are evaluation-only here. The lexicon derives from the English
Wiktionary dump via kaikki.org (CC BY-SA), acquired with dump date
and hash recorded.

Every component carries a row in a rights ledger in one of three
states (commercial-cleared, research-only, blocked-pending-counsel),
and runs fail closed if an uncleared component reaches a product
path. The released voice \cite{glornua2026} is the population voice
that passes the nearest-speaker anonymity test; the small-donor
blends that fail it are not released. Training audio is not
redistributed, per the terms of the corpus host.

\end{document}